%% file: main-camera.tex
\documentclass[letterpaper]{article} 
\usepackage{aaai2026}  
\usepackage{times}  
\usepackage{helvet}  
\usepackage{courier}  
\usepackage[hyphens]{url}  
\usepackage{graphicx} 
\usepackage{natbib}  
\usepackage{caption} 
\usepackage{algorithm}
\usepackage{algorithmic}
\usepackage[most]{tcolorbox}
\usepackage{colortbl}
\usepackage{soul}     
\usepackage{xcolor}   
\usepackage{multirow}
\usepackage{multicol}
\usepackage{booktabs}
\usepackage{graphicx}    
\usepackage{subcaption}  
\usepackage{float} 
\definecolor{grey}{gray}{0.85}

\usepackage{xcolor}

\newcommand*\colourcheck[1]{%
  \expandafter\newcommand\csname #1check\endcsname{\textcolor{#1}{\ding{52}}}%
}
\colourcheck{blue}
\colourcheck{green}
\colourcheck{red}
\usepackage{pifont}
\newcommand{\xmark}{\ding{55}}%

\usepackage{newfloat}
\usepackage{listings}
\DeclareCaptionStyle{ruled}{labelfont=normalfont,labelsep=colon,strut=off} 
\floatstyle{ruled}
\newfloat{listing}{tb}{lst}{}
\floatname{listing}{Listing}
\title{Concept‑RuleNet: Grounded Multi‑Agent Neurosymbolic Reasoning in Vision Language Models}
\author{
    Sanchit Sinha, Guangzhi Xiong, Zhenghao He, Aidong Zhang\\
}
\affiliations{
    University of Virginia, USA\\
    \url{{sanchit, guangzhi, zhenghao, aidong}@virginia.edu}
}

\usepackage{bibentry}

\begin{document}

\maketitle

\begin{abstract}
Modern vision-language models (VLMs) deliver impressive predictive accuracy yet offer little insight into `why' a decision is reached, frequently hallucinating facts, particularly when encountering out-of-distribution data.  Neurosymbolic frameworks address this by pairing black-box perception with interpretable symbolic reasoning, but current methods extract their symbols solely from task labels, leaving them weakly grounded in the underlying visual data. In this paper, we introduce a multi-agent system - \textbf{Concept-RuleNet} that reinstates visual grounding while retaining transparent reasoning. Specifically, a multimodal concept generator first mines discriminative visual concepts directly from a representative subset of training images. Next, these visual concepts are utilized to condition symbol discovery, anchoring the generations in real image statistics and mitigating label bias. Subsequently, symbols are composed into executable first-order rules by a large language model reasoner agent - yielding interpretable neurosymbolic rules. Finally, during inference, a vision verifier agent quantifies the degree of presence of each symbol and triggers rule execution in tandem with outputs of black-box neural models, predictions with explicit reasoning pathways. Experiments on five benchmarks, including two challenging medical-imaging tasks and three underrepresented natural-image datasets, show that our system augments state-of-the-art neurosymbolic baselines by an average of 5\% while also reducing the occurrence of hallucinated symbols in rules by up to 50\%. 
\end{abstract}

\input{src/introduction}
\input{src/related}
\input{src/methodology}

\input{src/experiments}

\input{src/conclusion}



\bibliography{aaai2026}

\appendix
\input{src/appendix}

\end{document}

%% file: src/introduction.tex
\section{Introduction}
\label{sec:intro}

\begin{figure}[t]
    \centering
    \includegraphics[width=0.49\textwidth]{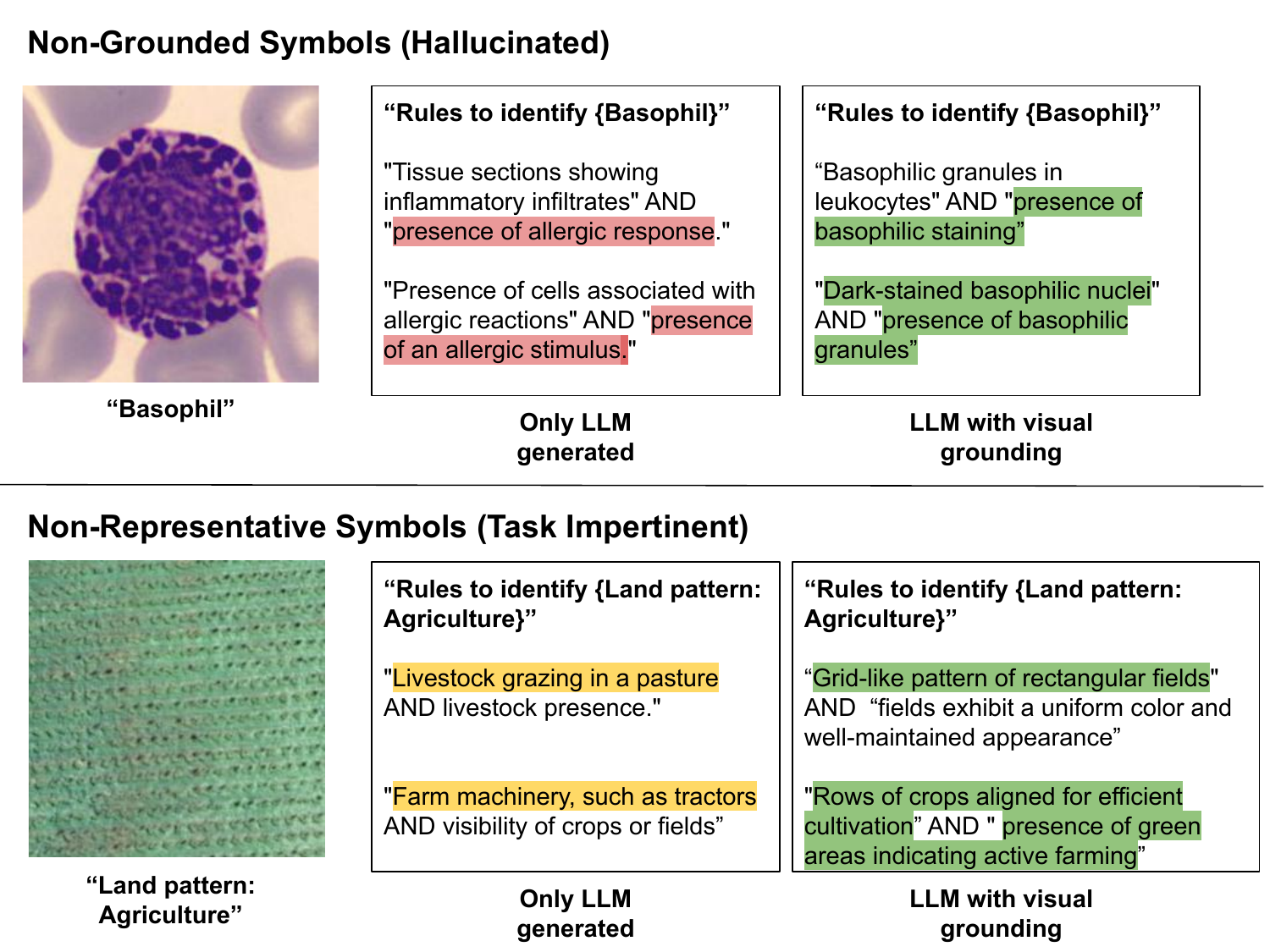}
    \caption{Examples sampled from the BloodMNIST \cite{yang2023medmnist} and UC-Merced Land Use \cite{yang2010bag} test datasets demonstrating a sample from the `Basophil' and `Agriculture Land Pattern' classes, respectively. We list the top rules influencing the decision-making process. (TOP) We observe that utilizing no images during the symbolic rule generation process (Symbol-LLM) generates rules with non-grounded symbols, i.e., symbols NOT present in test images (hallucinations). (BOTTOM) We observe that the generated rules are often somewhat semantically related to the task label but not representative of the task. The highlighted symbols are most relevant for prediction, with red symbols being hallucinated, green being appropriate, and yellow being non-representative.}
    \label{fig:exampl-nsai}
    \vspace{-15pt}
\end{figure}

With the increasing size and complexity of pre-trained large-scale Vision Language Models (VLMs) achieving widespread success in diverse vision tasks, it is tempting to utilize them for diverse use cases. Usually, VLMs are pre-trained on vast amounts of paired image-text data, which makes their learned decision-making process increasingly \textit{misaligned} with the human thought process. Researchers refer to such systems as \textbf{System-1} due to their speed, scalability, and unintuitive reasoning pathways behind a prediction. On the other hand, the human thought process is slower, more deliberate, and \textit{logical} - often composing multiple semantics \textit{neurosymbolically} to reach a more accurate and trustworthy decision - classified as \textbf{System-2} \cite{nye2021improving}. For example, consider the image in Figure~\ref{fig:exampl-nsai}, where both VLMs and logical rules output the same prediction, but System-2 reasoning is much more explainable and intuitive. Most of the current research on \textit{alignment problem} focuses on inducing System-2 reasoning in System-1 reasoners during pre-training, chain of thought \cite{wei2022chain}, etc., with limited success due to the fundamental assumptions geared towards scalability and efficiency. However, human cognition naturally weaves together System-1 (fast, associative) and System-2 (slow, deliberative) reasoning. Recognizing this, recent research has begun combining both systems to harness System-1's efficiency while using System-2 mechanisms to refine predictions and provide transparent, step-by-step rationales - thus leveraging both System-1 and System-2 \textit{in tandem}.

Popular works that combine System-1 and System-2 reasoning often utilize meticulously curated symbols to form neurosymbolic rules, and the prediction score of a rule is computed using \textit{First-order Logic}. For example, in \cite{yi2018neural}, the authors utilize several visual classifiers to select relevant functional tools that output the likelihood of a particular concept present in the image, which is then composed using a curated logical rule. Although such approaches can be an effective solution in a closed setting with limited rules and a closed set of concepts (e.g., in \cite{yi2018neural}, only four types of objects are considered), they are often not generalizable to large-scale complex datasets. As a consequence, a relatively new line of research proposes utilizing external knowledge from Large Language Models (LLMs) to automatically generate symbols (Generation) and subsequently construct logical rules using the generated symbols through neurosymbolic composition (Reasoning). For example, Symbol-LLM \cite{wu2024symbol} utilizes an LLM (GPT-3.5) first to extract all relevant symbols relating to human activity labels and then uses the same LLM as a reasoner for rules construction, overcoming expensive and slow manual logical rule generation.

Even though approaches like \cite{wu2024symbol} utilize LLMs as agents, they implicitly make a strong assumption that the parametric knowledge encoded in LLMs (during pre-training) is sufficient for effective symbol and rule generation. Note that the symbol discovery process in such approaches is conditioned \textit{only on a single task label}, with \textbf{no information} from the actual training images in the dataset. This presents a significant problem with datasets out-of-distribution to LLM pre-training, where the LLM parametric knowledge is lacking \cite{li2024improving,li2023multi}. Utilizing training images during the symbol generation and rule formation process provides multiple benefits, namely, grounding and representativeness as discussed below.

\noindent \textbf{Grounding:} Firstly, as LLMs are susceptible to hallucinations on contexts with limited knowledge \cite{simhi2025trust, zhang2024knowledge}, the symbols and rules generated can be adulterated with \textit{outright incorrect symbols} that are never encountered in the dataset. Consider the top example in Figure~\ref{fig:exampl-nsai}. Here, the symbols `presence of allergic response' and `presence of allergic stimulus' are encountered in neither the training set nor the test set - implying the symbol is hallucinated in context to this setting. However, conditioning symbol generation on training images adds semantic context during the generation process and is essential to ensure grounding.

\noindent \textbf{Representativeness:} Secondly, one of the main reasons why LLMs output irrelevant symbols for underrepresented tasks and surprisingly accurate symbols for a select few tasks is not due to their extensive parametric encoded knowledge, but due to the well-known phenomenon of \textit{dataset leakage} during pretraining \cite{carlini2021extracting} wherein LLMs are overfitted on descriptions of commonly utilized benchmark datasets. This phenomenon is subtle, but it is clear to observe in the results presented in \cite{wu2024symbol} and the similar work \cite{zhang2024large}, where performance on in-domain datasets is benchmarked, but on out-of-domain data is lacking. Similarly, consider the symbols generated for the `Land Pattern: Agriculture' class in Figure~\ref{fig:exampl-nsai} using \cite{wu2024symbol}, where the symbols are relevant to the description of the class but are irrelevant for the task at hand, i.e., recognizing land use patterns - making the symbols non-representative of the task.

As a consequence, in this paper, we propose \textbf{Concept-RuleNet} - a collaborative multi-agent framework which enforces \textbf{grounding} and \textbf{representativeness} in the neurosymbolic rule generation process. Concept-RuleNet - (i) effectively leverages a subset of training images to first extract grounded concepts using a \textbf{Visual Concept Extraction agent}, (ii) creates neurosymbolic symbols and composes them into logical rules using a strong \textbf{Symbol Exploration and Neurosymbolic agent}, and (iii) further augments standard System-1 prediction with symbolic predictions using a \textbf{Verifier agent}. With this three-agent system, we achieve the desired symbol properties by conditioning automatic symbol generation not only on the target labels but also grounded descriptions extracted as visual concepts from images in the training set. Subsequently, we demonstrate the drawbacks of current label-conditioned symbols and rule formation methods by empirically evaluating the improved prediction performance and degree of grounding compared to Concept-RuleNet. Lastly, we propose an extension, Concept-RuleNet++, which not only utilizes relevant symbols but also utilizes counterfactual symbols as a combination of conjunctive and disjunctive rule formation. More objectively, our contributions are as follows:
\begin{itemize}
    \item We propose Concept-RuleNet, a neuro-symbolic multi-agent system that utilizes three distinct LLM and VLM agents collaborating to extract grounded visual concepts and create representative and grounded symbols.
    \item We benchmark Concept-RuleNet across 5 challenging datasets on 4 modern VLMs and empirically demonstrate Concept-RuleNet's superior performance as compared to the state of the art (SOTA) approaches.
    \item We show that Concept-RuleNet produces \textbf{grounded} and \textbf{representative} symbols for more accurate rule generation and reduced hallucinations of LLM.
    \item We propose an extension - Concept-RuleNet++, which augments propositional rule generation in System-2 reasoning systems by leveraging \textbf{counterfactual} symbols for even higher prediction performance.
    
\end{itemize}


%% file: src/related.tex
\section{Related Work} 
\label{sec:related}

\noindent\textbf{Neurosymbolic Reasoning.} Neurosymbolic reasoning seeks to bridge the gap between the high-dimensional, often opaque representations learned by deep neural networks and the discrete, interpretable symbols fundamental to human reasoning \cite{besold2021neural}. Early works in this area utilized specialized architectures and regularization techniques to explain deep neural networks (DNNs) via propositional logic \cite{riegel2020logical,dong2019neural,garcez2023neurosymbolic}. Building on these foundations and informed by taxonomical frameworks such as that proposed by \citet{nye2021improving}, recent research has increasingly aimed to integrate fast, intuitive System‑1 processes with slower, deliberative System‑2 reasoning \cite{saha2024system, wu2024symbol,mao2019neuro}. Moreover, studies combining concept‑based explanations with neurosymbolic approaches have enhanced both interpretability and robustness \cite{barbiero2023interpretable}.

\medskip

\noindent\textbf{Utilizing Agents to Augment Black-box Models.} Multiple recent approaches leverage the extensive semantic knowledge encoded in large language models (LLMs) to directly generate meaningful symbolic representations and enhance the reasoning capabilities of VLMs and supplement their limited linguistic understanding \cite{chen2023large}. For example, methods such as those in \cite{oikarinen2023label} leverage the inherent language understanding of LLMs to extract meaningful symbols, while other works \cite{moayeri2023text, yang2023language} not only generate these concepts but also align them with visual data (VLMs). By harnessing both pre-training knowledge and the in-context learning capabilities of LLMs, these approaches are able to generate semantically rich symbols that serve as a bridge between raw visual inputs and higher-level reasoning tasks \cite{wu2024symbol,zhang2024large}. Approaches such as \cite{cho2023davidsonian, hu2023tifa} incorporate rich language cues into VLM inference through mechanisms like scene graphs or language priors, while other work \cite{zhou2023vicor} directly feeds LLM outputs into the visual understanding process. 

\medskip

\noindent\textbf{Comparisons to Related Work.} Our approach can be directly compared against agentic neurosymbolic systems, which leverage LLMs as symbol extractors and logical rule generators. We compare against Symbol-LLM \cite{wu2024symbol}, which relies solely on task labels as conditioning for generating symbols and rules, whereas our framework leverages visual concepts to induce grounding and representativeness. Even though Symbol-LLM achieves benchmark performance on HICO and Stanford datasets, the methodology is not generalizable to underrepresented or out-of-domain datasets. This is because HICO and Stanford borrow most of their samples from the bigger MSCOCO dataset and captions, a benchmark dataset used in most large-scale VLM pre-training approaches. As a consequence, Symbol-LLM is lacking in generating sound symbols when the dataset is out of domain to pre-training, as shown in Table 1 in \cite{zhang2024large} on `ALI' family of datasets and our experiments on datasets that are out-of-domain or underrepresented. 

%% file: src/methodology.tex


\begin{figure*}[ht]
  \centering
  \includegraphics[width=0.7\textwidth]{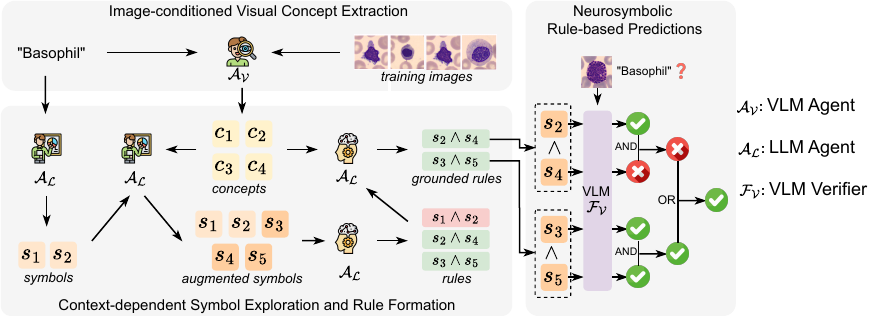}
  \caption{Schematic figure of Concept-RuleNet approach. Concept-RuleNet operates in three sequential stages - (i) Grounded Visual Concept Extraction: outputs visual concepts grounded in representative training images, (ii) Conditional Symbol Generation and Neurosymbolic Rule-based Predictions: explores relevant symbols and composes them into logical rules, and (iii) Neurosymbolic Rule-based Predictions: verifies the presence of each symbol in the rule to provide final predictions.}
  \vspace{-10pt}
  \label{fig:architecture}
\end{figure*}

\section{Methodology}
\label{sec:method}
In this section, we discuss the proposed Concept-RuleNet approach. We first begin by detailing the problem setting and formalizing notations. Next, we discuss the three primary stages of Concept-RuleNet, namely, Image-conditioned Visual Concept Extraction, Context-dependent Conditional Symbol Exploration and Rule Formation, and Neurosymbolic Rule-based Predictions. Finally, we discuss an extension to Concept-RuleNet - Concept-RuleNet++, which leverages counterfactual symbols to augment the neurosymbolic reasoning process. A schematic diagram of Concept-RuleNet is depicted in Figure~\ref{fig:architecture}. The subsequent inference process is depicted in Figure~\ref{fig:nsai-pred}.

\subsection{Preliminaries and Problem Setting}
\label{subsec:setting}
\noindent \textbf{System-1:} Let $\mathcal{X}$ be the space of input images and $\mathcal{Y}$ be the set of corresponding class labels. A typical System-1 model for image classification can be mathematically defined as a function mapping $F_{\text{sys1}}: \mathcal{X} \rightarrow \mathcal{Y}$ such that,
\[
y = F_{\text{sys1}}(x) ~~ \forall~(x,y)\in \{(\mathcal{X,Y})\}
\]
The function $F_{\text{sys1}}$ can be modeled as a neural network. Note that in this work, we primarily consider a \textit{zero-shot} setting where the System-1 model is used off the shelf, as fine-tuning large System-1 models is extremely expensive.

\noindent \textbf{System-2.} As opposed to learning a single function $F_{\text{sys1}}$, a System-2 model can be thought of as a composition of three separate functions $F_{concept}$, $F_{neurosymbolic}$, and $F_{verify}$ such that $F_{\text{sys2}} = F_{verify} \circ F_{neurosymbolic} \circ F_{concept}$. Note that $\circ$ represents function composition, i.e.,  $F_{neurosymbolic} \circ F_{concept}$ represents the output of $F_{concept}$ is input to $F_{neurosymbolic}$. More precisely, the function $F_{concept}$ maps $\mathcal{X} \rightarrow \mathcal{P(C)}$, where $\mathcal{P(C)}$ represents the power set of $\mathcal{C}$, a set consisting of relevant, human-understandable descriptive \textit{concepts}. Mathematically,
\begin{equation}
    c = F_{concept}(x),~\text{s.t.,}~ c  \subseteq \mathcal{C}
\end{equation}

Note that concepts $\mathcal{C}$ represent human-understandable descriptions of the images.  
Next, the function $F_{neurosymbolic}$ utilizes these visual concepts to explore task-relevant symbols, which are further composed into logical rules. Formally, $F_{neurosymbolic} : \mathcal{P}(\mathcal{C}) \rightarrow \mathcal{P}(\mathcal{L})$. Let $\mathcal{S}$ be the vocabulary of binary atomic symbols. A logical rule $l\in\mathcal{L}$ is formed by composing multiple symbols $s_i \in \mathcal{S}$. Mathematically,
\begin{equation}
    l = F_{neurosymbolic}(c),~\text{s.t.,}~ l \subseteq \mathcal{L}
\end{equation}
where each $l_i\in l$ is of the form $l_i = \bigwedge_{s_i \in s} s_i$. Finally, the function $F_{verify}$ maps $\mathcal{P(L)} \rightarrow \mathcal{Y}$ by implicitly scoring each symbol in a rule and then aggregating the scores into an entailment confidence, and returning the rule's prediction. Mathematically,
\begin{equation}
    F_{sys2} = F_{verify}(l), \text{where}~ l = F_{neurosymbolic} \circ F_{concept} (x)
\end{equation}

To leverage both System-1 and System-2 reasoning together in prediction, the final composite prediction is expressed by a weighted sum of System-1 and System-2 reasoning models:  
\begin{equation}
    \hat{y} = (1-\lambda) F_{sys1}(x) + \lambda F_{sys2}(x)
\end{equation}
where $\lambda$ controls the influence of System-2 model's prediction to the final prediction. In practice, all the individual functions of $F_{sys2}$, i.e., $F_{concept}$, $F_{neurosymbolic}$ and $F_{verify}$ are implemented using agents as discussed below.

\subsection{Image-conditioned Visual Concept Extraction}
As discussed, a System-2 reasoning model requires the generation of logical rules to emulate the human reasoning process. As discussed in Sec~\ref{subsec:setting}, the first stage is represented as the function $F_{concept}$ and generates a set of grounded and representative visual concepts. We utilize the VLM agent ($\mathcal{A_V}$) to first extract visual concepts present in each training image. Recent research has found that VLMs are extremely effective in identifying attributes in the images but less effective in identifying complex relationships between the discovered attributes, and hence act better as `Bag-of-visual attributes' than extracting complex relationships \cite{doveh2023teaching,herzig2023incorporating}. Hence, we extract low-level visual concepts $c_y \in \mathcal{P(C)}$ for each training image $x\in \mathcal{X}_{train}$ in the dataset, conditioned also on the task label $y$. Mathematically,
\begin{equation}
    c_y = \bigcup {\mathcal{A_{V}}(x_i,y,M)},~\forall~x_i~ \in \mathcal{X}_y^{\text{train}}
\end{equation}
where $\mathcal{A_V}$ is a function characterized by the task label $y$ and a number of concepts $M$, and $\mathcal{X}_y^{\text{train}}$ is the training subset of $\mathcal{X}$ with labels $y$. The visual concepts for randomly selected images belonging to a particular label $y$ in the training set are extracted and appended to a set, discarding duplicates. Finally, we get a set $c_y$ of observed visual concepts for each task label $y \in \mathcal{Y}$.

\subsection{Context-dependent Symbol Exploration and Rule Formation}
In the next stage, the function $F_{neurosymbolic}$ generates symbols and logical rules conditioned on the visual concepts. This stage has two distinct components - exploration and rule-formation, discussed below.

\noindent \textbf{Exploration.} To form grounded and representative symbols, we utilize a strong linguistic agent depicted by $\mathcal{A_{L}}$.  The process of symbol discovery is conditioned on both the task labels $y$ and the generated concepts $c_y$ in the last stage. To begin, we initialize the symbol set with dataset-specific symbols generated using an Initialization Symbol function ($\mathcal{IS}$), which is characterized by the task label $y$ and the number of initial symbols $K$. Mathematically, the symbol set is initialized as:
\begin{equation}
    S = \mathcal{A_{L}}(\mathcal{IS}(y,K) )
\end{equation}

Subsequently, we begin iterative rule exploration using both the initial symbols and the concepts associated with each label, as collected in the last stage. The visual concepts provide grounded context to $\mathcal{A_{L}}$ during exploration to minimize hallucinations. We utilize an Explore-Symbol function ($\mathcal{ES}$) characterized by the task label $y$ with context $c_y$.
Mathematically,
\begin{equation}
    S = S \bigcup \mathcal{A_{L}}(\mathcal{ES}(c_y,y))
\end{equation}
where the set $S$ collects all explored symbols following certain constraints as discussed in the next section.

\noindent \textbf{Rule Formation.} The next stage composes the symbols generated during the Exploration stage into logical neuro-symbolic rules through the linguistic agent $\mathcal{A_{L}}$ and an entailment function $\mathcal{EN}$. To construct a rule, we utilize the initialized symbol set $S$ as input, and each new symbol discovered during the explore stage is evaluated in the form of a rule in the Disjunctive Normal Form (DNF). Mathematically, a set of rules $l^*$ can be formed as:
\begin{equation}
    l^* = \{\bigwedge s_i \rightarrow y\}, \text{where~} s_i \in S~\text{and}~y \in \mathcal{Y}
\end{equation}
To ascertain the soundness of each rule in $l^*$, we utilize $\mathcal{A_{L}}$ to calculate entailment. Mathematically, it can be written as:
\begin{equation}
     l = \{~l_i\in l^*~~ |~~ \mathcal{A_L}(\mathcal{EN}(c_y,l_i)) > \epsilon \}
\end{equation}
where $\mathcal{A_L}$ calculates the entailment for a rule $l_i$ and $\epsilon$ is the entailment threshold signifying if the rule is plausible. Note that the rule scoring is dependent not just on the labels $y$, but also on the visual concepts extracted in the previous stage.

Based on the scoring mechanism, only the rules above a pre-defined threshold ($\epsilon$) are considered for System-2 reasoning. We limit the length of each rule to $N$ symbols, preventing overfitting a particular rule to the task. We point out that the functions $\mathcal{IS}$, $\mathcal{ES}$, and $\mathcal{EN}$ are modeled as prompt templates, which are detailed in the extended version.

\subsection{Neurosymbolic Rule-based Predictions}

For the final System-2 prediction, the process is decomposed into two steps: (i) computing scores for individual symbols using the \textit{verifier agent} ($\mathcal{F_{V}}$), and (ii) aggregating symbol scores to evaluate confidence of a rule.

For a neuro symbolic rule $l_i$ of the form $l_i = s_{i_1} \land s_{i_2} \land \cdots \land s_{i_k}$, where $\{s_{i_1}, s_{i_2}, \ldots, s_{i_k}\} \in \mathcal{P(S)}$ and a test image $x \in \mathcal{X}^{test}$, the overall score is then computed by taking the minimum of the scores of the individual symbols:
\begin{equation}
\begin{aligned}
    F_{verify}(l)  =  \max_{i\in\{1,\cdots,|l|\}}\{ & \min \{ \mathcal{F_V}(x, s_{i_1}),\\ \, & \mathcal{F_V}(x, s_{i_2}),\, \ldots,\, \mathcal{F_V}(x, s_{i_k}) \} \}.
    \label{eq:per-rule-verify}
\end{aligned}
\end{equation}
where $\mathcal{F_V}$ is a VLM, and the final System-2 prediction  over all rules in $\mathcal{L}$.

\begin{figure}[h]
    \centering
    \includegraphics[width=0.4\textwidth]{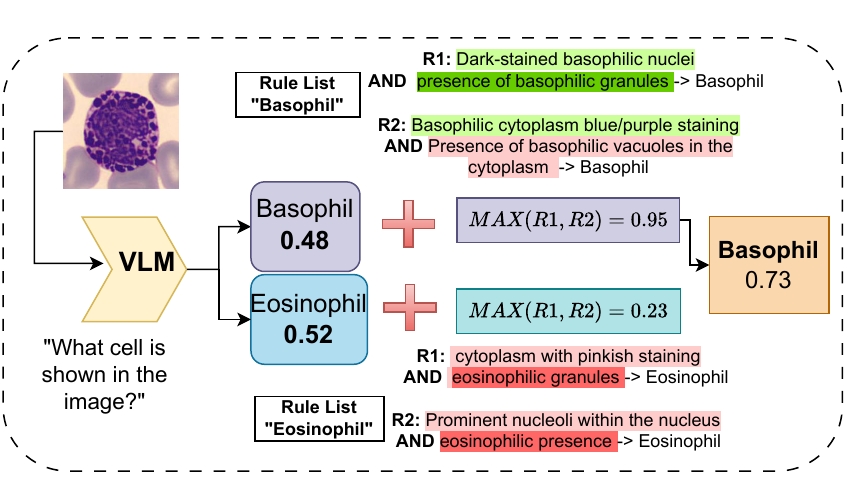}
    \vspace{-5pt}
    \caption{Inference process for a sample from the `Basophil' class from BloodMNIST dataset. VLM inference output assigns a probability score of $0.48$ to correct class. }
    \label{fig:nsai-pred}
    \vspace{-10pt}
\end{figure}

\noindent \textbf{Sample Inference Process.} An example of an inference procedure is demonstrated in Figure~\ref{fig:nsai-pred}. During inference, the image is passed through the System-1 model to infer the probability of each class label. (Basophil=0.48 and Eosinophil=0.52). In parallel, the Verifier Agent predicts the likelihood of each symbol for all neurosymbolic rules. For each class, the most likely rule is calculated using Equation~\ref{eq:per-rule-verify}. The final class prediction is performed through a weighted sum of System-1 (Basophil=0.48) and neurosymbolic output (Basophil=0.95). As can be seen, the neurosymbolic output `corrects' System-1 output to instill higher trust in predictions.
\vspace{-5pt}

\noindent \textbf{Theoretical intuition of grounding}: The Shannon entropy over symbols is $H(S)=\sum_{i} h(s_i),$ where $h(s_i)$ is the entropy of $s_i$. With grounding on image $x$, $H(S|x)=\sum_{i} h(s_i|x)$. With CRN, $H(S|x)<H(S)$ (\ref{fig:ground-exp}), indicating that grounding reduces uncertainty.

\subsection{Concept-RuleNet++}
Recent work has underscored the value of counterfactual reasoning in enhancing the interpretability of neurosymbolic systems. Counterfactual symbols serve as a complement to relevant symbols, for clearer decision boundaries and to reduce the impact of spurious correlations. Approaches like \cite{wachter2017counterfactual,dandekar2023counterfactual} empirically demonstrate that incorporating counterfactual symbols leads to improved generalization. We propose an extension to the standard Concept-RuleNet setup - Concept-RuleNet++ which augments reasoning with \textit{counterfactual} symbols. We utilize symbols present in rules from other classes with inverse verification, i.e., the probability of not being present. Mathematically, the new logical rules are formed as,
\begin{equation}
    l = \{\bigwedge \{\bigvee\{\Tilde{s}_i, s_i\}\}  \rightarrow y\}, 
\end{equation}
where $s_i, \Tilde{s_i} \in S$ and $y \in \mathcal{Y}$. $\Tilde{s_i}$ represents counterfactual symbols discovered. Note that Concept-RuleNet++ expands the rule structure to be a combination of both DNF and Conjunctive Normal Forms (CNF), forming a Mixed Normal Form structure of System-2 reasoning.

%% file: src/experiments.tex
\section{Experiments}
\label{sec:results}

\subsection{Dataset and Model Description}

\noindent\textbf{Dataset Description.} We utilize 5 medical and real-world datasets. \textbf{MedMNIST} \cite{yang2023medmnist} - BloodMNIST and DermaMNIST are designed for blood cell and skin abnormality classification, respectively. \textbf{UC-Merced Satellite Land Use} \cite{yang2010bag} and \textbf{WHU} \cite{xia2010WHURS19} are large-scale satellite image-based remote sensing datasets with high-resolution images for categorizing the land-use pattern. \textbf{iNaturalist-21} \cite{van2018inaturalist} consists of 13 classes referring to various biological species.
\subsubsection{System-1 Models.}
We benchmark our approach on 3 open-source VLMs - InstructBLIP-XXL \cite{dai2023instructblip}, LLaVA-1.5 \cite{liu2023llava}, and LLaVA-1.6 \cite{llava2024}. For MedMNIST, we also utilize a medical VLM - LLaVA-Med \cite{li2023llavamed}.

\noindent\textbf{Visual Concept Extraction and Symbol-Generation Models.}
For the MedMNIST family of datasets, we utilize LLaVA-Med as a visual concept extractor agent ($\mathcal{A_V}$) while for real-world datasets, we utilize LLaVA-1.6. The symbol exploration and rule formation agents are chosen to be strong LLMs with large-scale pre-training. We utilize \textbf{GPT-4o-mini} by OpenAI \cite{openai2024gpt4o} - a SOTA LLM with advanced reasoning capabilities.

\begin{table*}[ht]
\centering
\resizebox{0.98\textwidth}{!}{
\begin{tabular}{c*{5}{cccc}}
\toprule
 & \multicolumn{3}{c}{\textbf{BloodMNIST}} &  \multicolumn{3}{c}{\textbf{DermaMNIST}} & \multicolumn{3}{c}{\textbf{UCMerced-Satellite}} & \multicolumn{3}{c}{\textbf{WHU}} & \multicolumn{3}{c}{\textbf{iNaturalist}}  \\
\cmidrule(lr){2-4}\cmidrule(lr){5-7}\cmidrule(lr){8-10}\cmidrule(lr){11-13}\cmidrule(lr){14-16}
 & \textbf{S1} & \textbf{S-LLM} & \textbf{CRN} & \textbf{S1} & \textbf{S-LLM} & \textbf{CRN} & \textbf{S1} & \textbf{S-LLM} & \textbf{CRN} & \textbf{S1} & \textbf{S-LLM} & \textbf{CRN} & \textbf{S1} & \textbf{S-LLM} & \textbf{CRN}  \\
\hline
\textbf{Verifier:} Same as System-1   \\
\hline
\textbf{InstructBLIP}  & 11.55 & 13.56 & \textbf{18.09}  &  5.05 & 5.05 & \textbf{8.54} & 41.33 & 48.0  & \textbf{57.33} & 14.28  & 14.28 & \textbf{20.40} & 52.13 & 52.65 & \textbf{53.21}   \\
\textbf{LLaVA-1.5}  & 11.55 & 10.05 & \textbf{14.57}  & 9.54 & 30.15 &  \textbf{47.73}   & 65.33 & 64.00 & \textbf{69.33} & 18.36 & 18.36 & \textbf{19.22} & 58.12 & \textbf{60.24} & \textbf{60.24}\\
\textbf{LLaVA-1.6}  & 10.05 & 9.67 & \textbf{19.35} &  36.68 & 38.19 & \textbf{48.74}  & 38.66 &  49.33  &  \textbf{50.66}  & 27.27 & \textbf{28.43} & \textbf{28.43} & 61.30 & 61.30 & \textbf{63.45}  \\
\bottomrule
\end{tabular}}
\caption{Comparison between prediction accuracy on BloodMNIST, DermaMNIST, UCMerced-Satellite, WHU, and iNaturalist datasets across multiple VLMs. The columns under each dataset indicate System-1 (S1), S1 augmented with Symbol-LLM (S-LLM) and Concept-RuleNet (CRN), respectively.}
\label{tab:dataset-performance}
\vspace{-10pt}
\end{table*}

\subsection{Hyperparameter Settings}
\textbf{Visual Concept Extraction.} We utilize LLaVA-Med for extracting visual concepts in the MedMNIST dataset and LLaVA-1.6 for the other datasets with temperature 0.2. 

\noindent\textbf{Symbol Generation.} We utilize 5 initial premise symbols ($N$) followed by a maximum rule length of 3. To reduce overfitting, the rules on the visual context, we ensure maximum entailment ($\epsilon$) scores greater than 0.7 for a rule to be relevant. Increasing rule sizes beyond 3 provides diminishing returns (Refer Figure~\ref{fig:rule-length}). We run recursive symbol exploration and rule composition for 10 and 7 iterations, respectively, for MedMNIST and other datasets. The temperature is set at 0.7 for exploration and 0 for entailment. 

\noindent\textbf{Dataset Specific.} We utilize $\lambda=0.5$ for the Blood and Derma datasets while $\lambda=0.7,0.5,0.7$ for Satellite, WHU, and iNaturalist datasets respectively, based on tuning on the validation set.

\begin{table}[h]
\small
\centering
\resizebox{0.42\textwidth}{!}{
\begin{tabular}{c|ccc|ccc}
\toprule
 & \multicolumn{3}{c}{\textbf{BloodMNIST}} &  \multicolumn{3}{c}{\textbf{DermaMNIST}} \\
\cmidrule(lr){2-4}\cmidrule(lr){5-7}
\textbf{Model} & \textbf{S1} & \textbf{S-LLM} & \textbf{CRN} & \textbf{S1} & \textbf{S-LLM} & \textbf{CRN} \\
\midrule
\textbf{InstructBLIP}  & 11.55 & 11.55  & \textbf{13.56}  & 5.05 & 5.05 & \textbf{7.86}  \\
\textbf{LLaVA-1.5}  & 11.55 & \textbf{13.56} & \textbf{13.56}  & 9.54  & 30.15 &  \textbf{34.21} \\
\textbf{LLaVA-1.6}  & 10.05 & 9.54 & \textbf{13.21} & 36.68 & 48.25 &  \textbf{66.33}   \\
\textbf{LLaVA-Med} & 11.05  & 11.05  & \textbf{12.06}  & 4.81 & 2.77 &  \textbf{12.56}   \\
\bottomrule
\end{tabular}}
\caption{Prediction accuracy for BloodMNIST and DermaMNIST datasets with a medical verifier agent - LLaVA-Med. We observe that LLaVA-Med boosts performance significantly on the DermaMNIST dataset.}
\label{tab:med-verifier}
\vspace{-5pt}
\end{table}

\subsection{Implementation Details}
\noindent \textbf{Baseline Replication.} We recreate the Symbol-LLM baseline \cite{wu2024symbol} by adapting it to different datasets. Note that Symbol-LLM is exclusively tested on Human Activity Recognition (HOI) datasets (which, as discussed before, is in-domain to LLM pre-training), hence its efficacy on the selected datasets is unknown. We utilize the implementation with minor changes in prompts by changing task labels corresponding to the datasets used in this paper. We limit the rule lengths to 3 symbols each, with a lowered minimum entailment value of 0.7 (to achieve speed up). As demonstrated in \cite{wu2024symbol}, rules with at most 3 symbols are functionally equivalent to longer rules.

\noindent \textbf{Verification.} Although VLMs are adept at generating natural language descriptions of input images, the actual token predictions are underexplored. We augment VQAScore \cite{lin2025evaluating} scoring strategy by reformatting the visual description into a binary `Yes/No' question. Mathematically, for a given $x \in \mathcal{X}^{test}$, the prediction probability of a symbol $s$ for a VLM ($F_V$) with logit outputs ($\hat{F}_{V})$,
\begin{equation}
    P(\text{``yes''}|x,s) = e^{\hat{F}_{V}  [\text{``yes''}]}/({e^{\hat{F}_{V}[\text{``yes''}]} + e^{\hat{F}_{V}[\text{``no''}]})}
\end{equation}
The probability of the token `Yes' is taken as the proxy for the \textit{confidence} of prediction.

\subsection{Experiment-1: Prediction Performance}
\noindent \textbf{Choice of Verifier.} As the verifier ($\mathcal{A_V}$) is one of the most important aspects of System-2 models, we consider 2 real-world settings - one where all the symbols and rules are pre-computed \textit{ante-hoc} and only the rules and test-images are available. In this case, the System-1 model can act as a verifier with minor modifications. The next setting is where we are provided both test images and an \textit{inference} only verifier. In this case, to provide more confident symbol probabilities, we can utilize a domain-specific verifier. We chose LLaVA-Med, a strong medical VLM, as a verifier.

\noindent \textbf{Medical Datasets.}
We report the prediction performance of Concept-RuleNet as compared to baseline Symbol-LLM and only without any System-2 integrations (S1) in Table~\ref{tab:dataset-performance}. We observe that Concept-RuleNet consistently outperforms System-1 only and Symbol-LLM baselines on most datasets. For the BloodMNIST dataset and DermaMNIST datasets, Concept-RuleNet beats Symbol-LLM by an average of about 5\% across all model settings.
\\
\noindent \textbf{Real-world Datasets.}
Here, we observe that System-1 models demonstrate good performance out of the box on the Satellite datasets - UCMerced and WHU as they share resemble with types of images encountered in the pre-training setup of VLMs. We observe that Concept-RuleNet outperforms Symbol-LLM by a considerable margin of about 5\% on most models, with the highest improvement being on the Instruct-BLIP-XXL model of 9.33\% on the UCMerced dataset, while an average of 2-4\% on the WHU dataset, possibly due to the more challenging nature of the images. Finally, the prediction performance of Concept-RuleNet is superior to Symbol-LLM on the iNaturalist dataset on all models. The results are a testament to our approach, as it improves prediction performance on underrepresented, out-of-distribution datasets.

\noindent \textbf{Utilizing domain-specific verifiers.} Next, we observe that using a medical verifier improves performance in both medical datasets in Table~\ref{tab:med-verifier}, making a strong case to consider designing even more powerful medical VLMs as verifiers in the future. Lastly, in Row-4, we conduct an interesting experiment where we utilize LLaVA-Med as both a System-1 model and a verifier, which interestingly does not give good results. This is an important insight - implying that models like LLaVA-Med do not have a deep understanding of the images (weak reasoning) but can be good verifiers. 

\noindent\textbf{Improvements using Concept-RuleNet++.}
Next in Table~\ref{tab:Concept-RuleNet-++}, we report the performance improvement using Concept-RuleNet++. Note as Concept-RuleNet++ utilizes counterfactual symbols, it is not directly comparable to Symbol-LLM but rather an approach similar to combining Symbol-LLM with counterfactual symbols which is out of scope for this work. We observe Concept-RuleNet++ outperforms Concept-RuleNet by an average of 1-2\%, making it ideal for use cases that are performance-sensitive.
\begin{table}[ht]
\centering
\resizebox{0.35\textwidth}{!}{
\begin{tabular}{ccc}
\toprule
\textbf{Dataset} & \textbf{Concept-RuleNet} & \textbf{Concept-RuleNet++} \\
\midrule
BloodMNIST & 18.09 & \textbf{21.43} \\
DermaMNIST & 8.54 & \textbf{14.23}\\
\midrule
Satellite & 57.33 & \textbf{58.12}  \\
WHU & 20.40 & \textbf{21.52} \\
iNaturalist & 53.21 & \textbf{54.15}\\
\bottomrule
\end{tabular}}
\caption{Prediction performance improvements for Concept-RuleNet++ over Concept-RuleNet. We observe a consistent improvement in prediction performance over all datasets except iNaturalist, possibly due to the extreme diversity in the training samples, as the occurrence of a counterfactual symbol is still pretty high. (All datasets tested on InstructBLIP)}
\label{tab:Concept-RuleNet-++}
\vspace{-10pt}
\end{table}

\begin{table}[ht]
\centering
\resizebox{0.36\textwidth}{!}{
\begin{tabular}{cccc}
\toprule
\textbf{Initialization} & \textbf{Exploration} & \textbf{Entailment} & \textbf{Prediction} \\
\midrule
\xmark & \xmark & \xmark & 48.00 \\
\redcheck & \xmark & \xmark & 49.50 \\
\redcheck & \redcheck & \xmark & 55.10 \\
\redcheck & \redcheck & \redcheck & 57.33 \\
\bottomrule
\end{tabular}}
\caption{Ablation study for presence of visual context in each stage. First row corresponds to Symbol-LLM setting.}
\label{tab:ablation}
\vspace{-15pt}
\end{table}

\subsection{Experiment-2: Symbol Quality}
\begin{figure}[h]
    \centering
    \begin{subfigure}[t]{0.22\textwidth}   
        \centering
        \includegraphics[width=1.1\linewidth]{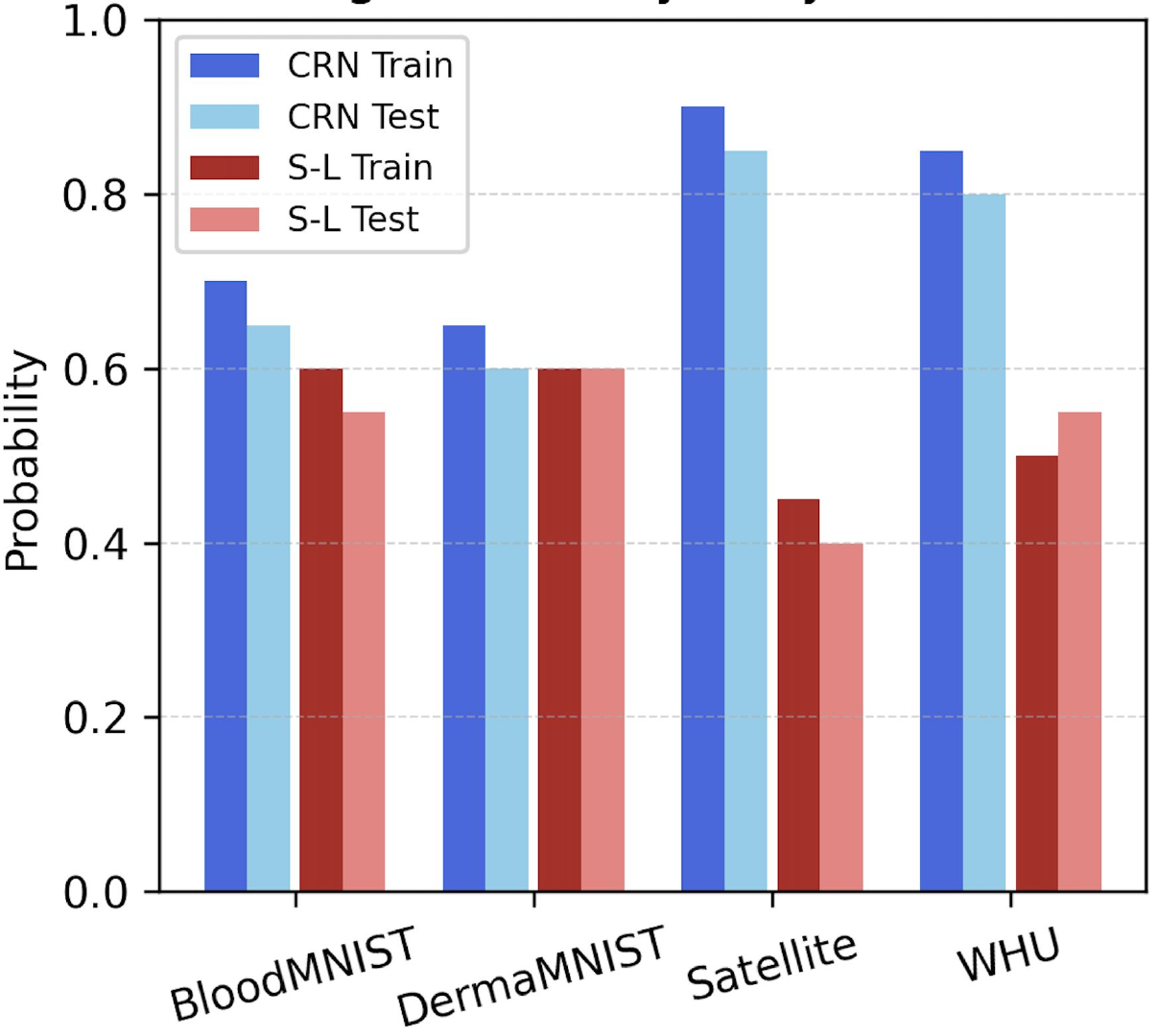}
        \caption{Degree of grounding on symbols generated by Concept‑RuleNet (CRN) and Symbol‑LLM (S-LLM) using the InstructBLIP model.}
        \label{fig:ground-exp}
    \end{subfigure}%
    \hfill
    \begin{subfigure}[t]{0.22\textwidth}
        \centering
        \includegraphics[width=1.25\linewidth]{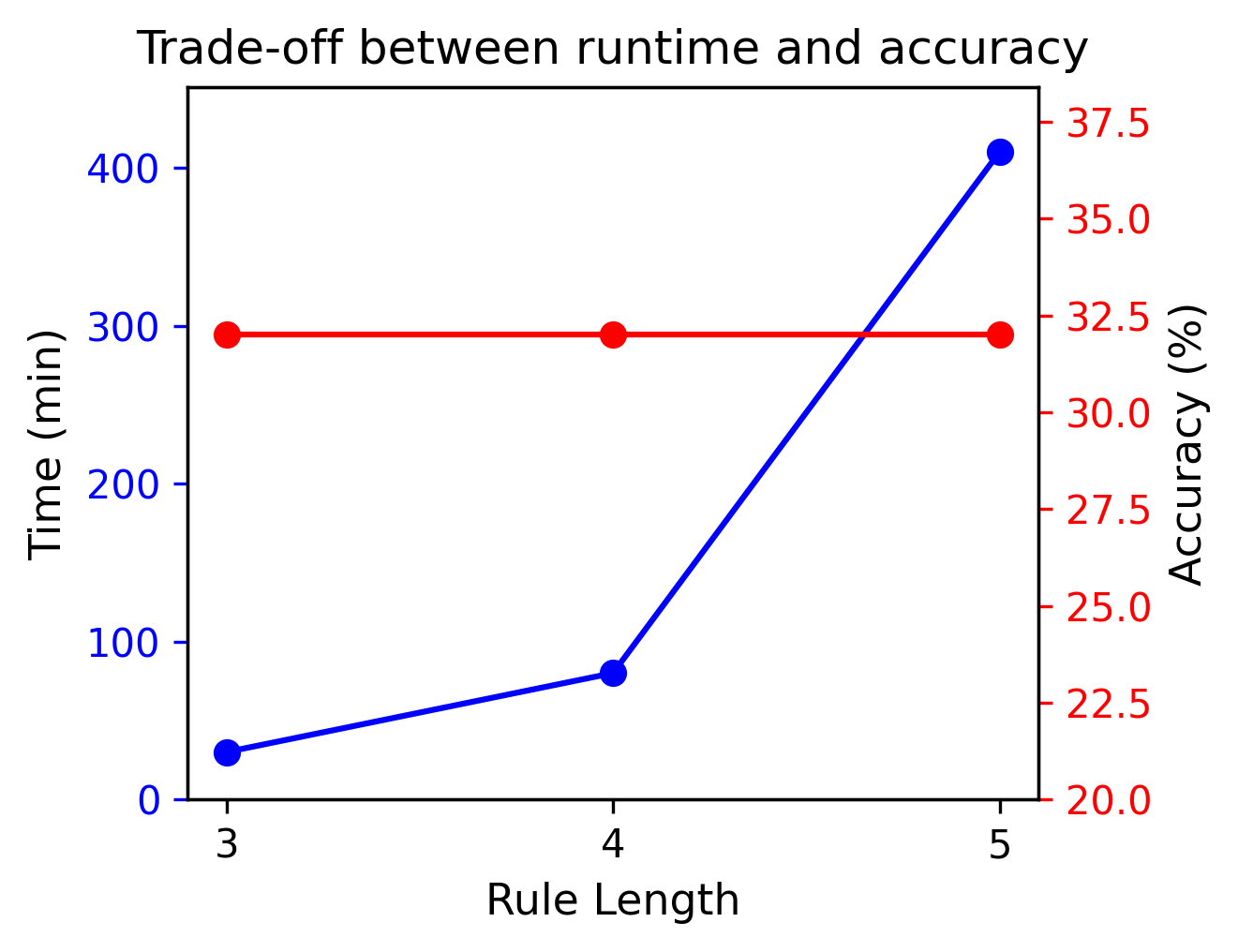}
        \caption{Trade‑off between runtime and accuracy on a subset of the UCMerced‑Satellite data. Longer rules yield diminishing accuracy gains.}
        \label{fig:rule-length}
    \end{subfigure}
    \vspace{-5pt}
    \caption{(a) Symbol grounding and (b) rule length–accuracy tradeoff.}
    \label{fig:grounding-and-tradeoff}
    \vspace{-10pt}
\end{figure}

\noindent \textbf{Quantitative Grounding Measures.} To validate if the symbols generated by Concept-RuleNet are better grounded, we compute the average likelihood of each symbol in the generated rules being present in both the train and test images using a VLM as shown in Figure~\ref{fig:ground-exp}. We observe that symbols generated by Concept-RuleNet are more likely to be present in both the Train and Test sets than those by Symbol-LLM. For Satellite and WHU datasets, the difference is even more stark, where Symbol-LLM's symbol occurrence rate is less than 0.5 - highlighting the need for Concept-RuleNet in underrepresented domains.

\noindent \textbf{Representativeness of Symbols.} Similarly, to evaluate the representativeness of generated symbols we format them as a question, `How likely are \{symbol1, symbol2,..\} in predicting \{class\} for a \{task\}?' and pass them through an advanced reasoning model. We find that the average likelihood outputted for symbols generated by Concept-RuleNet is \textbf{0.54} as compared to 0.49 for Symbol-LLM. (Refer to the extended version for experiment design).

\subsection{Ablation Study}
\noindent \textbf{Impact of Visual Concepts.} We conduct an extensive ablation study by providing visual concepts as a context in each stage of the symbol generation and entailment process. Visual context improves each stage of Concept-RuleNet for the UC-Merced dataset, highlighting its usefulness.

\noindent \textbf{Impact of hyperparameters.} We further report the impact of $\lambda$, which controls the impact of System-2 reasoning on the final output and the initial number of images used for visual concept extraction in Table~\ref{tab:ablation-hp}. We observe that too high or too low $\lambda$ degrades performance. Similarly, considering too many images is detrimental due to overfitting on obscure, irrelevant concepts.

\begin{table}[h]
\centering
\resizebox{0.42\textwidth}{!}{%
\begin{tabular}{c|ccc|cc}
\hline
\textbf{Method} & \multicolumn{3}{c|}{$\mathbf{\lambda}$} & \multicolumn{2}{c}{\textbf{\# of images}}  \\ 
\hline
& 0.3 & 0.7 & 0.9 & 50  & 90 \\ 
\hline
\textbf{Symbol-LLM}  & 44.00 & 48.00 & 38.66 & 48.00* & 48.00*\\ 
\textbf{Concept-RuleNet}  & 49.33 & 57.33 & 48.00 & 57.33 & 56.28\\ 
\hline
\end{tabular}}
\caption{Ablation Study on the impact of hyperparameter $\lambda$ and number of images used for visual concept extraction. (*) implies no training images are utilized.}
\label{tab:ablation-hp}
\vspace{-10pt}
\end{table}

\noindent \textbf{Complexity analysis.} The time complexity of visual concept extraction depends on the number of images selected. Assuming the number of images is $n$ and the number of classes is $c$, the time complexity is given as $\mathcal{O}(nc)$. No other change in complexity between CRN and S-LLM. 


\begin{figure}[h]
    \centering
    \includegraphics[width=0.5\textwidth]{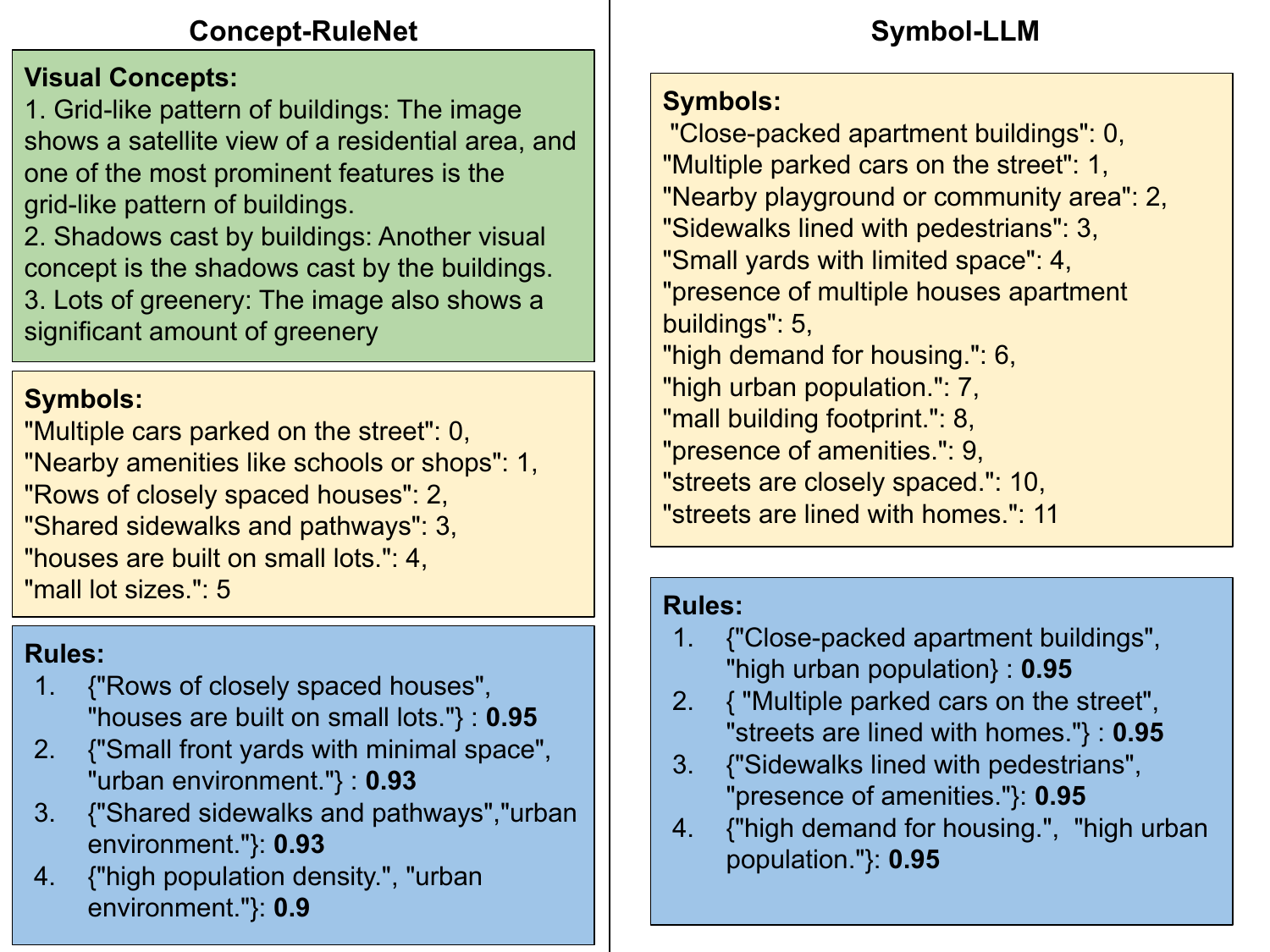}
    \caption{(LEFT) Concept-RuleNet generated visual concepts, symbols, and rules. (RIGHT) Symbol-LLM generated symbols and rules for the `denseresidential' class in the satellite dataset. We observe that Symbol-LLM outputs multiple symbols with high probability of being presented (represented as bold numbers in the rules), but which are non-representative of the task - classifying images into `denseresidential' category. E.g. `high-demand for housing' is an irrelevant symbol for this task.}
    \label{fig:eg-sat}
    \vspace{-10pt}
\end{figure}

%% file: src/conclusion.tex
\section{Conclusion}
\label{sec:conclusion}

In this paper, we propose Concept-RuleNet, a novel image-conditioned neurosymbolic reasoning framework designed to improve image classification performance. By leveraging both target labels and training images to generate grounded and representative symbols, our approach effectively mitigates the issues of hallucination and dataset leakage that have limited prior methods relying solely on label-conditioned symbol generation. The empirical evaluations across a diverse set of benchmark datasets demonstrate Concept-RuleNet's superior performance. Furthermore, we introduced Concept-RuleNet++, an extension that incorporates counterfactual symbols into the logical rule formation process. Overall, our work underscores the importance of integrating visual context into the neurosymbolic reasoning process and opens up promising avenues for future research aimed at developing more interpretable systems.

%% file: src/appendix.tex
\newpage
\clearpage
\section{Appendix}
\label{sec:appendix}

\noindent \textbf{Code:} The code for Concept-RuleNet can be found at: \url{https://github.com/sanchit97/Concept-RuleNet}

\subsection{Why System-2 Models over Fine-tuning?}
\textbf{Lack of Generalization:} Improving zero-shot performance of System-1 models is crucial because it removes the dependence on large-scale data curation and computationally expensive fine-tuning efforts. While fine-tuning can adapt a model to specific domains or tasks, it requires labeled data, often in significant quantities, and may still not generalize well outside the training distribution. In contrast, a strong zero-shot model can make predictions on novel tasks or underrepresented data domains without requiring any additional supervision, enabling rapid deployment to new applications, especially in resource-constrained or evolving scenarios. 

\textbf{Lack of interpretability:} zero-shot models generally rely on large pretrained networks whose internal reasoning remains unclear - leading to concerns about unintended bias, reliability, and user trust. By strengthening the zero-shot capabilities through grounded logical rules, we can reduce reliance on extensive labeled data while also introducing mechanisms to clarify or justify predictions. 
\subsection{Datasets}
\begin{itemize}
    \item \textbf{MedMNIST} \cite{yang2023medmnist} is designed for medical image classification. It comprises several sub-datasets, each focusing on a different medical imaging domain.  We shortlist two separate standalone datasets in themselves, for evaluation -  Blood (individual normal cells) and Dermatology (Dermatoscope images). We sample 200 images from each class for extraction.
    \item \textbf{UC-Merced Satellite Land Use \cite{yang2010bag} and WHU \cite{xia2010WHURS19}: } are large-scale satellite image-based remote sensing datasets with high-resolution images for categorizing the land-use pattern. The land use patterns belong to one of 21 different classes such as `agriculture', `beach', etc. for UC-Merced and 19 similar classes for WHU. We sample 50 images from each class for visual concept extraction with an unchanged test set.
    \item \textbf{iNaturalist-21} \cite{van2018inaturalist} consists of 13 classes representing supercategories of each species (Animalia, Mammalia, etc.). We sample 2000 images from the training set to be utilized as conditioning for visual concept extraction and sample, with a maximum of 100 per class. Finally, 50 images for each class are used from test set.
\end{itemize}
\vspace{-5pt}

\subsection{Model Settings}
\subsubsection{System-1 Models}
\begin{itemize}
\item \textbf{InstructBLIP-Flan-T5} \cite{dai2023instructblip} is a multimodal AI model designed for vision-language tasks, integrating the BLIP-2 \cite{li2023blip} framework with Flan-T5-XXL variants - a powerful text-to-text transformer from Google's Flan-T5 \cite{chung2022scaling}. 
\item \textbf{LLaVA-1.5-7B} \cite{liu2024improved} is an advanced vision-language model (VLM) that integrates LLaMA \cite{dubey2024llama}  with a visual encoder for multimodal understanding. It uses improved visual encoders based on CLIP \cite{radford2021learning} and instruction tuning to generate more context-aware and detailed responses. 
\item \textbf{LLaVA-1.6 (LLaVA-Next)-7B} \cite{llava2024} is the next iteration of the LLaVA-1.5, utilizing a stronger image encoder and diverse multimodal training data. LLaVA-Next uses an instruction-tuned LLM framework built on Llama-Vicuna \cite{vicuna2023} model.
\item \textbf{LLaVA-Med} \cite{li2023llavamed} is a version of the LLaVA-Next open-source VLM which has been pre-trained explicitly on diverse medical data, including clinical reports, medical images, pathologies, etc. (only for medical datasets)

\end{itemize}

\begin{table*}[t]
  \centering
\small
\begin{tabular}{lccccccc}
\toprule
\textbf{Comparison} & $\Delta$ (pp) & SD & $t(4)$ & $p$ (2-tail) & 95\% CI & Cohen’s $d$ & Sig.\@ \\
\midrule
CRN vs.\ S-LLM & \textbf{+4.99} & 2.95 & 3.79 & 0.019 & [\,1.3, 8.7\,] & 1.69 & Yes \\
CRN vs.\ S1    & \textbf{+6.83} & 5.45 & 2.80 & 0.048 & [\,0.1, 13.6\,] & 1.25 & Yes \\
\bottomrule
\end{tabular}
\caption{Paired two-sided $t$-tests across the five datasets using InstructBLIP as verifier.  
$\Delta$ is the mean difference in percentage points (pp) between CRN and S1.  
Sig. indicates significance at $\alpha = 0.05$.}
\label{tab:crn_vs_sllm_stats}
\end{table*}

\subsection{Implementing Concept-RuleNet}
In this section, we expand on the precise implementation details of Concept-RuleNet. For Stage-1, we utilize $\mathcal{A_{V}}$ - a strong VLM to extract visual concepts. The prompt is formatted as follows:
\begin{tcolorbox}[colback=grey,colframe=black,title=Prompt for Extracting Visual Concepts using $\mathcal{A_{V}}$]
In this picture, we see \{$label$\}. List $\{N\}$ visual concepts that can be seen in relation to \{$label$\}. 
\end{tcolorbox}

For the symbol initialization stage, we utilize a strong LLM as the function $\mathcal{A_L}$. The prompt structure for the initialization prompt $\mathcal{IS}$ is as follows:
\begin{tcolorbox}[colback=grey,colframe=black,title=Prompt template for $\mathcal{IS}$]
In a picture, we see $\{label\}$. List $\{k\}$ entities that can be seen that verify \{$label$\}. 
\end{tcolorbox}
Note that the context symbols for the prompt are prepended to the structure. Next, for the symbol exploration stage, we utilize the $\mathcal{EE}$ structure as detailed below.

\begin{tcolorbox}[colback=grey,colframe=black,title=Prompt template for $\mathcal{ES}$]
We know that for \{y\}, we generally observe \{$c_y$\}. Based on this, in a picture, if $\{$sym$\}$ AND [CONDITION] THEN \{$y$\}. What is [CONDITION]?  
\end{tcolorbox}

\newpage
This prompt structure allows us to explore one symbol every iteration. As more and more symbols are explored, we verify if they are relevant for task prediction by composing them into rules. The entailment scores are calculated using the entailment $\mathcal{EN}$ function. The prompt structure is as follows:

\begin{tcolorbox}[colback=grey,colframe=black,title=Prompt template for $\mathcal{EN}$ ]
We know \{$c_y$\} is responsible for \{$y$\}. Given \{$l$\}, how likely is \{$y$\}? Choose from the following options - (A) 0.1, (B) 0.5, (C) 0.7 (D) 0.9, (E) 0.95. 
\end{tcolorbox}

The neurosymbolic rules are considered relevant when the average entailment score is above a threshold $\epsilon$.

\subsection{Verifier Prompt template}
We utilize a standard prompt template for the verifier to assign probability scores for each symbol in a rule.

\begin{tcolorbox}[colback=grey,colframe=black,title=Prompt template for $\mathcal{A_V}$ as part of $F_{verify}$]
``In the image we can see a \{task\}. Does this image show \{symbol\}? Answer in Yes or No. "
\end{tcolorbox}
The final output is calculated as the softmax of the logits associated to the `yes' and `no' symbols.
\begin{equation}
    P(\text{``yes''}|x,s) = \frac{e^{\hat{F}_{V}[\text{``yes''}]}}{e^{\hat{F}_{V}[\text{``yes''}]} + e^{\hat{F}_{V}[\text{``no''}]}}
\end{equation}

\subsection{Concept-RuleNet++}
\begin{figure}[h]
    \centering
    \includegraphics[width=0.95\linewidth]{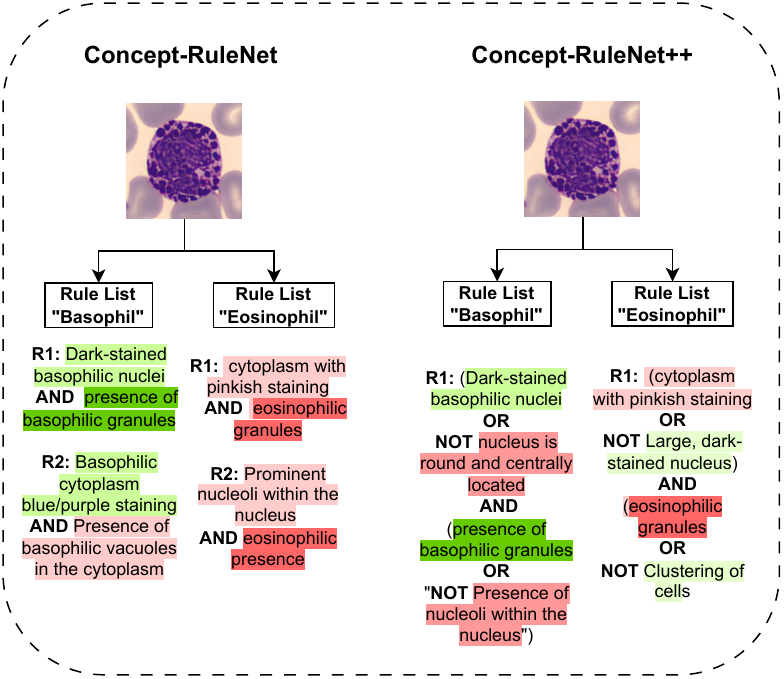}
    \caption{Concept-RuleNet vs Concept-RuleNet++'s System-2 Reasoning process on the sample from the Basophil class.}
    \label{fig:eg-sym-vlm-++}
\end{figure}

We describe the working of Concept-RuleNet++ by reconsidering the example figure. We randomly select symbols from other classes and add them as symbols to the existing rules as shown in Figure~\ref{fig:eg-sym-vlm-++}. In the figure, Basophil class rules is augmented with non-relevant symbols taken at random (for instance, a lymphocyte in this case) which is `nucleus is round and central' as shown in Figure~\ref{fig:eg-sym-vlm-++}. Adding a counterfactual symbol thus augments the rule reasoning process for more accurate predictions. Note that during entailment calculation, the probability of the `not present' symbol is taken as the inverse of its present probability, i.e. 
\begin{equation}
    P(\text{``no''}|x,\tilde{s}) = \frac{e^{\hat{F}_{V}[\text{``no''}]}}{e^{\hat{F}_{V}[\text{``yes''}]} + e^{\hat{F}_{V}[\text{``no''}]}}
\end{equation}

\subsection{Statistical Tests}
In Table~\ref{tab:crn_vs_sllm_stats}, we demonstrate the two-sided paired t-test values demonstrating the improvement of Concept-RuleNet over Symbol-LLM. All p-values are comfortably above 0.05, implying statistical significance of improvements.

\subsection{Rule-length and Diminishing Returns}
\begin{figure}[h]
    \centering
    \includegraphics[width=0.4\textwidth]{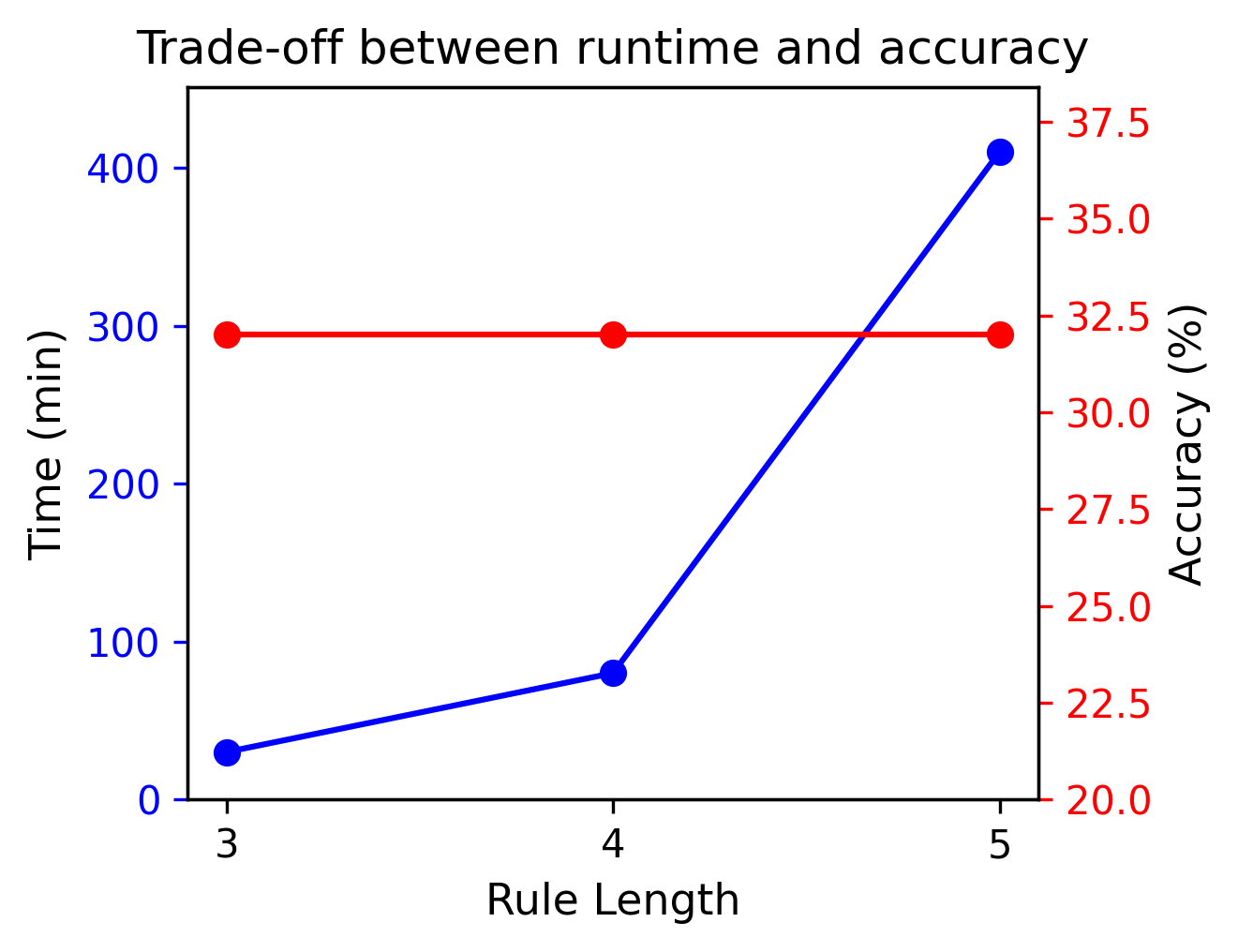}
    \caption{Tradeoff between runtime and accuracy}
    \label{fig:tradeoff-appendix}
\end{figure}

We limit the length of rules to 3 symbols each for efficiency during the rule generation process. In Figure~\ref{fig:tradeoff-appendix}, we show an actual rule prediction performance and time taken to compute the said rule, as a function of rule length. For the `agriculture' class, forming the rules on the OpenAI API increases exponentially with rule size with minimal improvement in performance for that particular class (Model used: InstructBLIP-XXL on the Satellite dataset).

\subsection{Representativeness Experiment Design}
To find out how representative the extracted symbols are for a particular task, we utilize the following prompt template for a strong reasoner - GPT-o1, the best reasoning LLM to our knowledge. We average over randomly chosen 3 class rules.

\begin{tcolorbox}[colback=grey,colframe=black,title=Prompt template for representativeness]
‘How likely are \{symbol1, symbol2,..\} in predicting \{class\} for a \{task\}? Output only a single probability value 
\end{tcolorbox}

We observe that Concept-RuleNet average is \textbf{0.54} as compared to 0.49 for Symbol-LLM, where we can understand that the symbols in the rules are more representative.

    \begin{figure*}[t]
  \centering
  \begin{subfigure}[b]{0.69\textwidth}
    \centering
    \includegraphics[width=\textwidth]{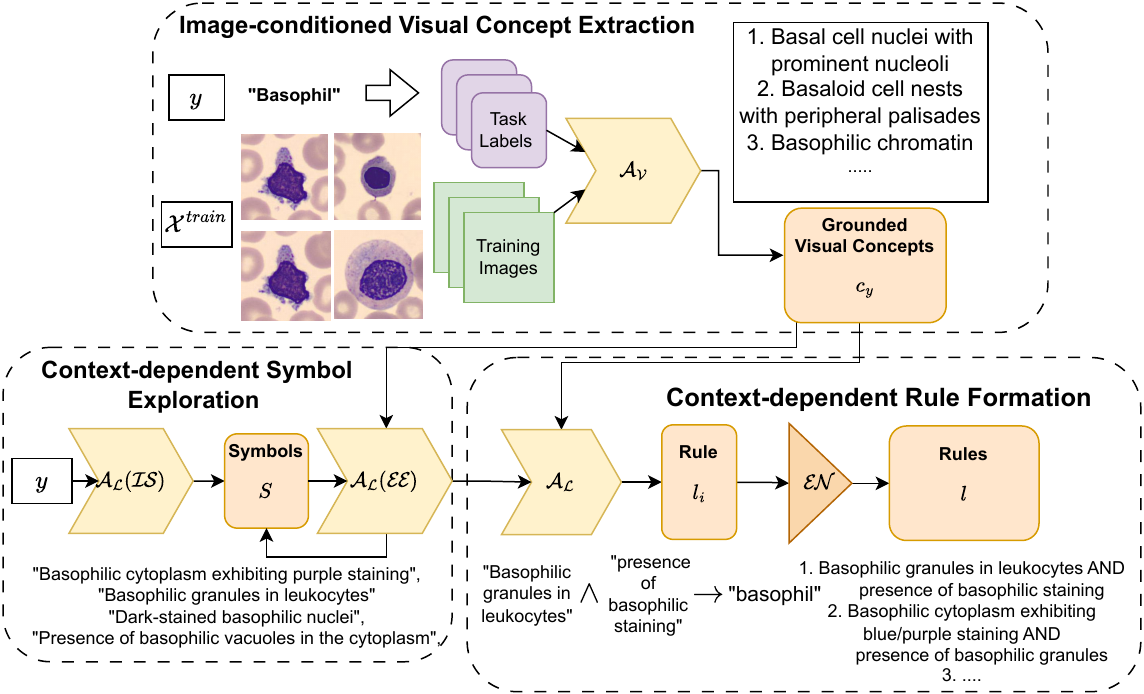}
    \caption{The proposed Concept-RuleNet approach in flowchart for grounded and representative symbol generation and subsequent rule formation.}
    \label{fig:schematic-symbol-vlm}
  \end{subfigure}
  \hfill
  \begin{subfigure}[b]{0.30\textwidth}
    \centering
    \includegraphics[width=\textwidth]{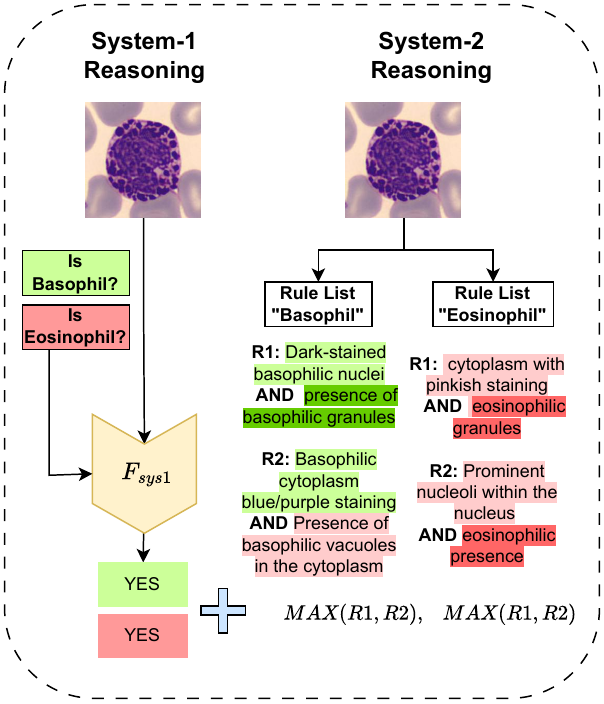}
    \caption{Schematic figure demonstrating integration of System-1 and System-2 reasoning process during inference. }
    \label{fig:system-2}
  \end{subfigure}
  \caption{Flow-chart based rule construction process. (a) Concept-RuleNet during training consists of 3 distinct stages - Image-conditioned Visual Concept Extraction, Context-dependent Symbol Exploration and Rule Formation. (b) For a given image, the System-1 model outputs the probability of each class directly using a fast and efficient model ($F_{sys1}$). Parallely, System-2 reasoning assigns probability scores for each symbol of each rule associated with each class. Finally, the rules with maximum probability give the final prediction probability for each class. Note the colors Green and Red represent probability values greater or less than the chosen boundary of classification (0.5) respectively with the color gradient representing the magnitude of difference from the boundary.}
  \label{fig:schematic}
  \vspace{-5pt}
\end{figure*}

\subsection{Example Concepts, Symbols, and Rules}
\begin{figure*}
    \centering
    \includegraphics[width=0.8\textwidth]{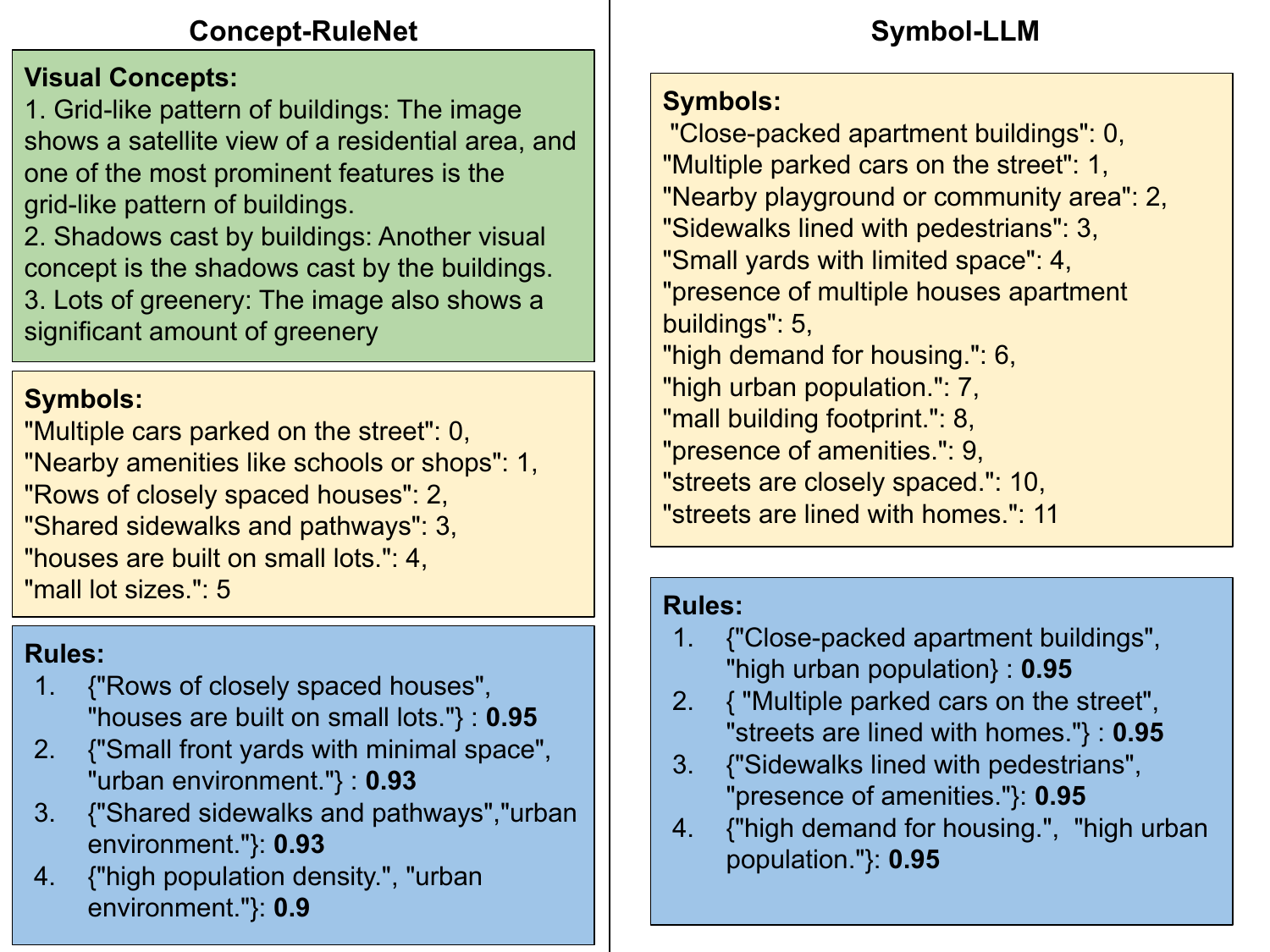}
    \caption{(LEFT) SConcept-RuleNet generated visual concepts, symbols and rules. (RIGHT) Symbol-LLM generated symbols and rules for the `denseresidential' class in the UCMerced satellite dataset.}
    \label{fig:eg-sat}
    
\end{figure*}

\begin{figure*}[t]
    \centering  
    \includegraphics[width=0.8\textwidth]{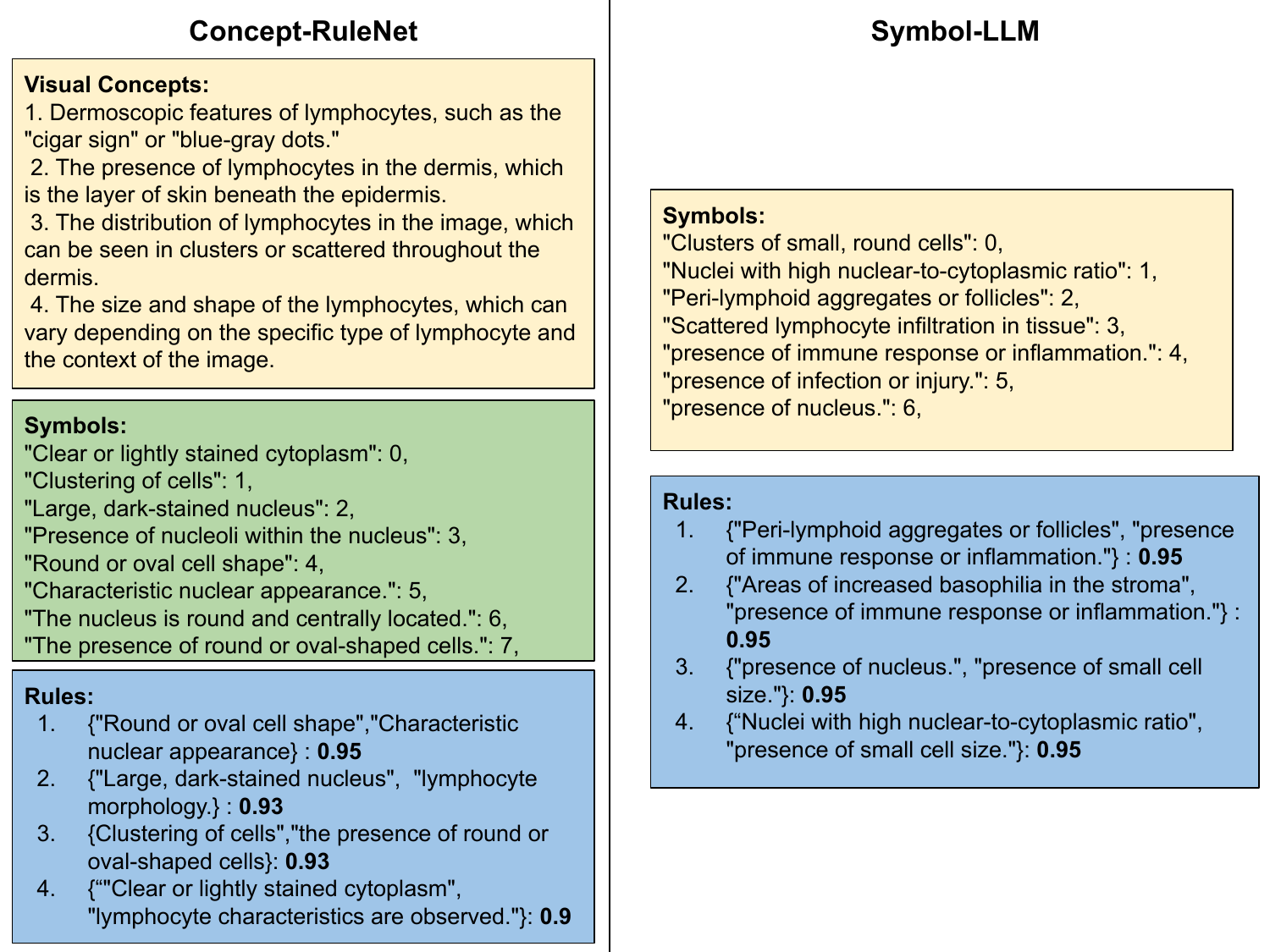}
    \caption{(LEFT) Concept-RuleNet generated visual concepts, symbols, and rules. (RIGHT) Symbol-LLM generated symbols and rules for the `lymphocyte' class in the BloodMNIST dataset.}
    \label{fig:eg-blood}
\end{figure*}

We demonstrate some example visual concepts, symbols and rules for BloodMNIST in Figure~\ref{fig:eg-blood} and Satellite in Figure~\ref{fig:eg-sat}. As can be observed, Concept-RuleNet learns more grounded and representative as compared to Symbol-LLM.


